\documentclass[runningheads]{llncs}

\usepackage[T1]{fontenc}
\usepackage{graphicx}
\usepackage{subcaption}
\usepackage{adjustbox}

\usepackage{hhline}

\usepackage{amsmath}
\usepackage{amssymb}

\begin{document}
\title{SqueezerFaceNet: Reducing a Small Face Recognition CNN Even More Via Filter Pruning 
}
%
%
\author{Fernando Alonso-Fernandez\inst{1} \and
Kevin Hernandez-Diaz\inst{1} \and \\
Jose Maria Buades Rubio\inst{2} \and
Josef Bigun\inst{1}}
\authorrunning{F. Alonso-Fernandez et al.}
%
\institute{School of Information Technology, Halmstad University, Sweden \\ \email{feralo@hh.se, kevin.hernandez-diaz@hh.se, josef.bigun@hh.se} \and Computer Graphics and Vision and AI Group, University of Balearic Islands, Spain \email{josemaria.buades@uib.es}}
\maketitle              
\begin{abstract}

The widespread use of mobile devices for various digital services has created a need for reliable and real-time person authentication.
%
%
In this context, facial recognition technologies have emerged as a dependable method for verifying users due to the prevalence of cameras in mobile devices and their integration into everyday applications. 
%
%
The rapid advancement of deep Convolutional Neural Networks (CNNs) has led to numerous face verification architectures. However, these models are often large and impractical for mobile applications, reaching sizes of hundreds of megabytes with millions of parameters. 
%
%
We address this issue by developing SqueezerFaceNet, a light face recognition network which less than 1M parameters. This is achieved by applying a network pruning method based on Taylor scores, where filters with small importance scores are removed iteratively.
Starting from an already small network (of 1.24M) based on SqueezeNet, we show that it can be further reduced (up to 40\%) without an appreciable loss in performance.
To the best of our knowledge, we are the first to evaluate network pruning methods for the task of face recognition. 

\end{abstract}

\section{Introduction}

The widespread use of smartphones as all-in-one platforms has led to more people relying on them for accessing online services such as e-commerce and banking. 
%
%
This makes it crucial to implement robust user authentication mechanisms to ensure secure device unlocking and protected transactions.
Here, we address face recognition (FR) for mobile applications, where biometric verification is increasingly employed for the mentioned purposes.
As with many other vision tasks, Convolutional Neural Networks (CNNs) have become a very popular tool for biometrics, including FR \cite{[Sundararajan18-DLbiometrics]}.
Nevertheless, the high-performing models proposed in the literature, e.g. \cite{Deng19CVPR_ArcFace}, usually entail extensive storage and computational resources due to their millions of parameters.
This poses a significant challenge for deploying them on resource-limited devices.

\begin{table*}[htb]

\caption{Proposed lightweight models in the literature for face recognition. 
}

\begin{center}

\begin{tabular}{|c|c|c|c|c|}


\hline

& Input & 
Para- & Vector & Base 
\\

Network & size & 
meters & Size& Architecture 
\\

\hline


LightCNN (18) \cite{[Wu18lightCNN]} & 128$\times$128 & 
12.6M & 256 &   
\\ \hline

MobileFaceNets (18) \cite{[Chen18MobileFaceNets]} & 112$\times$112 & 
0.99M & 256 &  MobileNetv2 
\\ \hline

MobiFace (19) \cite{[Duong19MobiFace]} & 112$\times$112 & 
n/a & 512 &  MobileNetv2 
\\ \hline

ShuffleFaceNet (19) \cite{[Martinez19ShuffleFaceNet]} & 112$\times$112 & 
0.5-4.5M & 128 &  ShuffleNet  
\\ \hline

SeesawFaceNets (19) \cite{[zhang19seesawfacenets]} & 112$\times$112 & 
1.3M & 512 & 
\\ \hline


VarGFaceNet (19) \cite{Yan2019ICCVW_VarGFaceNet} & 112$\times$112 &  
5M & 512 & VarGNet 
\\ \hline

SqueezeFacePoseNet \cite{[Alonso20SqueezeFacePoseNet]} & 113$\times$113 & 
0.86-1.24M & 1000 & SqueezeNet 
\\ \hline

PocketNet (21) \cite{boutros2021Access_Pocketnet} & 112$\times$112 &  
0.92-0.99   M & 128-256 & PocketNet 
\\ \hline

MixFaceNets (21) \cite{Boutros2021IJCB_MixFaceNets} & 112$\times$112 & 
1.04-3.95M & 512 & MixNets 
\\ \hline

\textbf{SqueezerFaceNet (ours)} & 113$\times$113 & 
0.65-0.94M & 1000 & SqueezeNet  
\\ \hline

\end{tabular}

\end{center}

\label{tab:networks}
\end{table*}
\normalsize

Across the years, several light CNNs have been presented, mainly for common visual tasks in the context of the ImageNet challenge \cite{[Russakovsky15_ImagenetChallege]}. 
Examples include 
SqueezeNet \cite{[Iandola16SqueezeNet]} (1.24M parameters), 
MobileNetV2 \cite{[Sandler18mobilenetv2]} (3.5M), 
ShuffleNet \cite{[Zhang18ShuffleNet]} (1.4M), 
MixNets \cite{Tan19BMVC_MixNets} (5M),
or VarGNet \cite{zhang2020vargnet} (13.23M). 
They employ different techniques to achieve fewer parameters and faster processing, such as point-wise convolution, depth-wise separable convolution, variable group convolution, mixed convolution, channel shuffle, and bottleneck layers.
Some works (Table~\ref{tab:networks}) have adapted these networks for FR purposes \cite{[Chen18MobileFaceNets],[Duong19MobiFace],[Martinez19ShuffleFaceNet],Yan2019ICCVW_VarGFaceNet,[Alonso20SqueezeFacePoseNet],Boutros2021IJCB_MixFaceNets}.
Instead of adapting existing common architectures, the work \cite{boutros2021Access_Pocketnet} suggested applying Neural Architecture Search (NAS) to design a family of light FR models, named PocketNets.

In this paper, we follow another strategy, consisting of applying network compression to existing architectures. 
Common techniques include knowledge distillation, quantization, or pruning.
A number of them have been used to reduce the size of general image classification models, and some works recently started to apply them for face detection \cite{Jiang22CCS_PruneFaceDet} or ocular recognition \cite{Rattani23ACCESS_OcularCNNPruningBenchmark}.
Here, we use a pruning method based on importance scores of network filters \cite{Molchanov19CVPRcnnPruningTaylor} to reduce an already small FR network of 1.24M parameters that uses a modified SqueezeNet architecture \cite{[Iandola16SqueezeNet],[Alonso20SqueezeFacePoseNet]}.
Thus, we call our network SqueezerFaceNet.
The importance score of a filter is obtained considering its effect on the error if it is removed. This is computed by first-order Taylor approximation, which only requires the elements of the gradient computed during training via backpropagation.

To the best of our knowledge, we are the first to evaluate network pruning methods for the task of FR.
We test SqueezerFaceNet on a face verification scenario over VGGFace2-Pose, a subset of the VGGFace2 database \cite{[Cao18vggface2]} with 11040 images from 368 subjects on three poses (frontal, three-quarter, and profile).
We show that the number of filters of the network can be reduced up to 15\% without a significant loss of accuracy in one-to-one comparisons for any given pose.
If we allow five images per user to build an identity template, accuracy is not significantly affected until a 30-40\% reduction in the number of filters.
%
%

\section{Network Pruning Method}
\label{sect:pruning_method}

We apply the method of \cite{Molchanov19CVPRcnnPruningTaylor}, 
which iteratively estimates the importance scores of individual elements 
based on their effect on the network loss. 
Then, elements with the lowest scores are pruned, leading to a more compact network.

Given a network with parameters 
$\textbf{W}=\left\{ w_{0},w_{1},...,w_{M} \right\}$ and a training set $\mathcal{D}$ of input $\left( x_{i} \right)$ and output $\left( y_{i} \right)$ pairs
$\mathcal{D}=\left\{ \left( x_{0},y_{0} \right),\left( x_{1},y_{1} \right),...,\left( x_{K},y_{K} \right) \right\}$, 
the aim of network training is to minimize the classification error $E$ by solving $\mathop {\min }\limits_{\textbf{W}} E(\mathcal{D},{\textbf{W}}) = \mathop {\min }\limits_{\textbf{W}} E(y|x,{\textbf{W}})$. 
%
%
The importance of a parameter $w_{m}$ can be defined by its impact on the error if it is removed.
Under an \textit{i.i.d.} assumption, the induced error can be quantified as the squared difference of the prediction error $E$ with and without the parameter:

\begin{equation}
\label{eq:induced_error}
    {\mathcal{I}}_m=\bigg( E\left( \mathcal{D},\textbf{W} ) - E( \mathcal{D},\textbf{W}|w_{m}=0 \right)  \bigg)^{2}
\end{equation}
    
However, computing $\mathcal{I}_m$ for each parameter using Eq.~\ref{eq:induced_error} would demand to evaluate $M$ versions of the network, one for each removed parameter, making the process expensive computationally.
This is avoided by approximating $\mathcal{I}_m$ in the vicinity of $\textbf{W}$ by its first-order Taylor expansion ${\mathcal{I}}_{m}^{1}(\textbf{W})=\left( g_{m}w_{m} \right)^{2}$, where $g_{m}=\frac{\partial E}{\partial w_{m}}$ are the elements of the gradient $g$. A second-order expansion is also proposed \cite{Molchanov19CVPRcnnPruningTaylor}, but it demands computing the Hessian of $E$, so we employ the first-order approximation for a more compact and fast computation.
The gradient $g$ is available from backpropagation, so $\mathcal{I}_m$ can be easily computed. 
To compute the joint importance of a set of parameters $\textbf{W}_S$ (e.g. a filter), we apply:

\begin{equation}
\label{eq:joint_induced_error}
    {\mathcal{I}}_{S}^{1}(\textbf{W}) \triangleq \sum_{s\in S}^{}\left( g_{s}w_{s} \right)^{2}
\end{equation}

The algorithm starts with a trained network, which is pruned iteratively over the same training set.
Given a mini-batch, the gradients are computed, and the network weights are updated by gradient descent.
Simultaneously, the importance of each filter is computed via Eq.~\ref{eq:joint_induced_error}.
At the end of each epoch, the importance scores of each filter are averaged over the mini-batches, and the filters with the smallest importance scores are removed.
The pruning process is then stopped after a certain number of epochs. 
The resulting network can be then fine-tuned again over the training set to regain potential accuracy losses due to filter removal.

\begin{figure}[htb]
\centering
        \includegraphics[width=.8\linewidth]{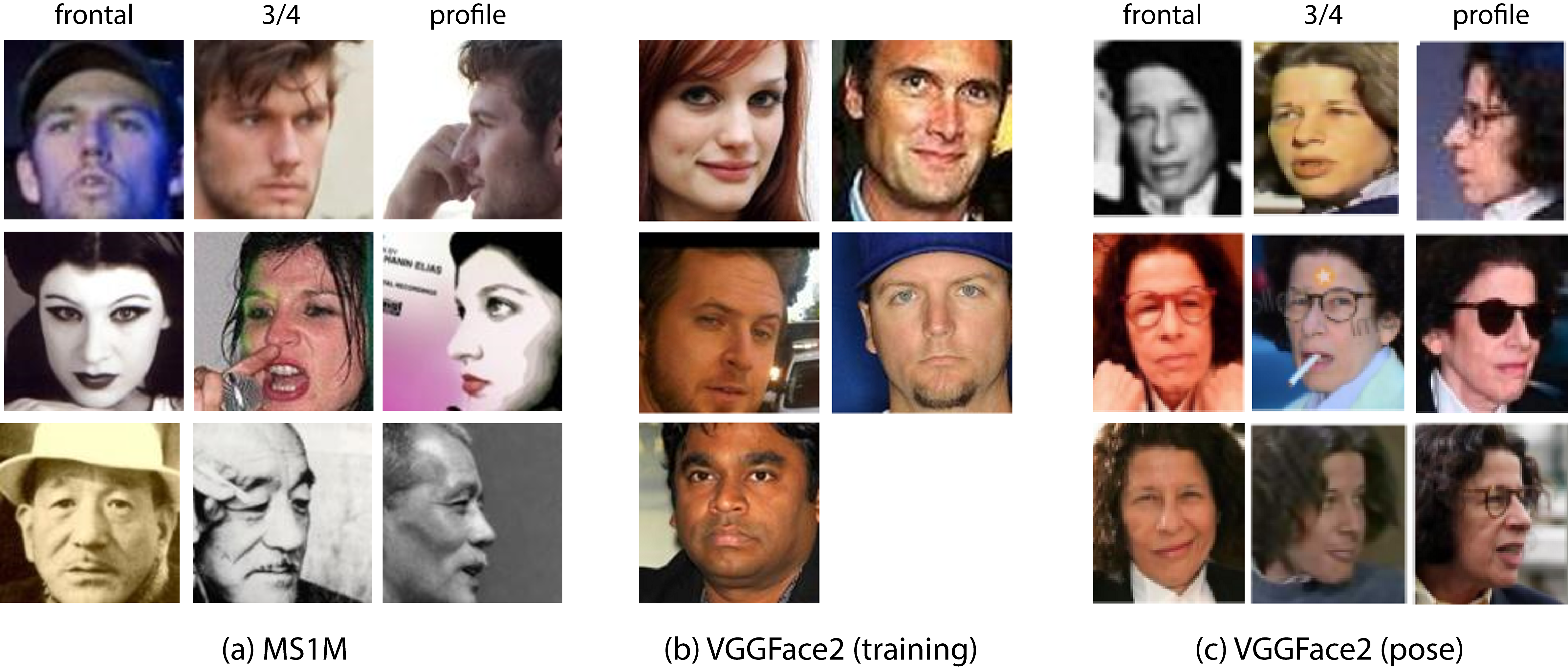}
\caption{Example images of the databases used. (a) MS1M from three users (by row) and three viewpoints (column). (b) VGGFace2 training images with a random crop. (c) VGGFace2 pose templates from 
three viewpoints (by column).
}
\label{fig:databases}
\end{figure}

\section{SqueezerFaceNet Architecture and Database}
\label{sect:network_db}

As the backbone for SqueezerFaceNet, we employ SqueezeNet \cite{[Iandola16SqueezeNet]}. 
This is among the smallest generic CNNs proposed in the context of the ImageNet challenge, and one of the early networks designed to reduce the number of parameters and size.
It has only 18 convolutional layers, 1.24M parameters, and 4.6 MB in its uncompressed version. 
%
%
%
To reduce the network size, it uses fire modules, which first reduce the input channel dimensionality via 1$\times$1 point-wise filters (\textit{squeeze} phase), to be then processed with a larger amount of (more costly) 3$\times$3 and 1$\times$1 filters in a lower dimensional space (\textit{expand} phase). 
%
%
Another strategy is late downsampling, so convolution layers are presented maps as large as possible. According to its authors, it should lead to higher accuracy.

In the present paper, we adopt the SqueezeNet implementation previously proposed for FR using light CNNs in \cite{[Alonso20SqueezeFacePoseNet]}, referred to as SqueezeFacePoseNet.
In particular, the network employs an input of 113$\times$113, instead of the original 227$\times$227 of SqueezeNet. This is achieved by changing the stride of the first convolutional layer from 2 to 1, while keeping the rest of the network unchanged, which allows to reuse ImageNet parameters as starting model. Such transfer learning strategy from ImageNet has been shown to provide equal or better performance than if initialized from scratch, while converging faster \cite{[Kornblith19imagenet_transfer_better]}.
In the present paper, we have also added batch normalization between convolutions and ReLU layers. 
This is missing in the original SqueezeNet and in \cite{[Alonso20SqueezeFacePoseNet]}, but batch normalization is commonly used before non-linearities to aid in the training of deep networks \cite{Ioffe15icmlBatchNorm}. 
Compared to \cite{[Alonso20SqueezeFacePoseNet]}, we observe that it also leads to increased recognition accuracy, with a small overhead of parameters.

The database for training and evaluation is VGGFace2, with 3.31M images of 9131 celebrities (363.6 images/person on average) \cite{[Cao18vggface2]}.
The images, downloaded from the Internet, show significant variations in pose, age, ethnicity,
lightning and background.
The protocol contemplates 8631 training classes (3.14M images) and the remaining 500 classes for testing.
For cross-pose experiments, a subset of 368 subjects from the test set is defined (called VGGFace2-Pose), having 10 images per pose (frontal, three-quarter, and profile) and a total of 11040 images.

To further improve recognition performance, we also pre-train SqueezerFaceNet in the RetinaFace cleaned set of the MS-Celeb-1M database \cite{[Guo16_MSCeleb1M]} (MS1M for short), with 5.1M images of 93.4K identities.
The release contains 113$\times$113 images of MS1M cropped with the five facial landmarks provided by RetinaFace \cite{[Deng19RetinaFace]}. 
While MS1M has a more significant number of images,
its intra-identity variation is limited due to an average of 81 images/person.
Following previous research \cite{[Cao18vggface2],[Alonso20SqueezeFacePoseNet]}, we first pre-train SqueezerFaceNet on a dataset with a large number of images (MS1M) and then fine-tune it with more intra-class diversity (VGGFace2). This has been shown to provide better performance than training the models only with VGGFace2. 
Some example images are shown in Figure \ref{fig:databases}.

\section{Experimental Protocol}
\label{sect:protocol}

SqueezerFaceNet is trained for biometric identification using the soft-max function and ImageNet as initialization.
We follow the training/evaluation protocol of VGGFace2 \cite{[Cao18vggface2]}.
For training, the bounding boxes of VGGFace2 images are resized, so the shorter side has 129 pixels, and a random crop of 113$\times$113 is taken. 
A random crop is not possible with MS1M, since images are directly at 113$\times$113.
We also apply horizontal random flip to both databases.
The optimizer is SGDM with a mini-batch of 128. The initial learning rate is 0.01, which is decreased to 0.005, 0.001, and 0.0001 when the validation loss plateaus. 
Two percent of images per user in the training set are set aside for validation.
Users in MS1M with fewer than 70 images are removed to reduce the parameters of the fully connected layer dedicated to under-represented classes and ensure at least one image per user in the validation set. This results in 35016 users and 3.16M images.
%
%
We train with Matlab r2022b and use the ImageNet pre-trained model that comes with such release.

\begin{figure}[htb]
\centering
\centering
\includegraphics[width=.45\textwidth]{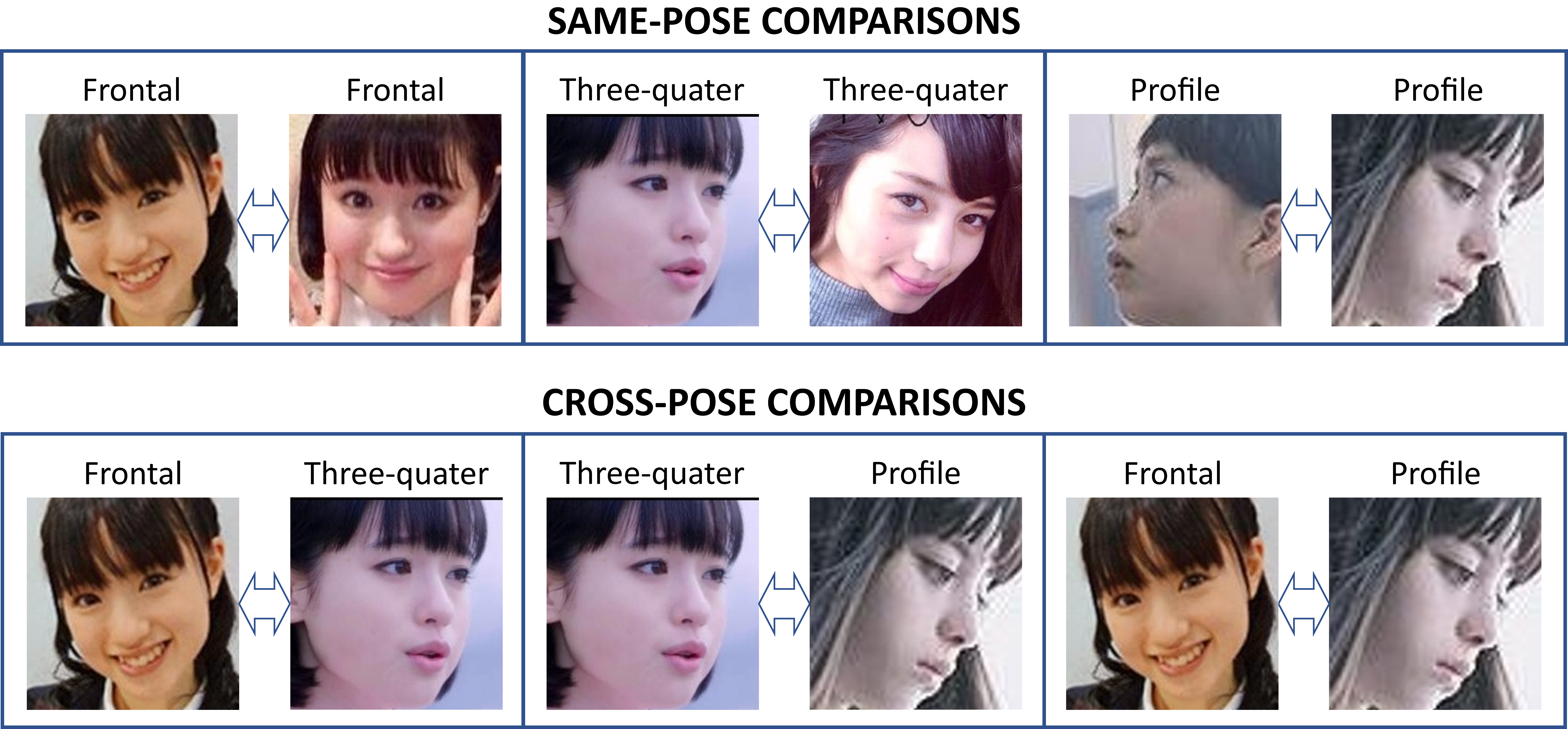}
\caption{Same-pose (top) and cross-pose comparisons (bottom). 
}
\label{fig:protocol}
\end{figure}

Verification experiments are done with VGGFace2-Pose following the protocol of \cite{[Cao18vggface2]}.
%
%
A center crop of 113$\times$113 is taken after the shortest image side is resized to 129 pixels.
%
%
Identity templates per user are created by combining five faces with the same pose, resulting in two templates available per user and pose. 
To test the robustness of the network and the pruning method in more adverse conditions, we also do experiments using only one image as template.
A template vector is created by averaging the descriptors of the faces in the template set, which are obtained from the layer adjacent to the classification layer (i.e., the Global Average Pooling). With SqueezeNet, this corresponds to a descriptor of 1000 elements.
To further improve performance against pose variation, we also average the descriptor of an image and its horizontally flipped counterpart, which is hypothesized to help to minimize the effect of pose variation \cite{[Duong19MobiFace]}. 
The cosine similarity is then used to compare two given templates.

\begin{table}[htb]

\caption{Face verification results on the VGGFace2-Pose database (EER \%) without pruning. F=Frontal View. 3/4= Three-Quarter. P=Profile.}

\centering

\begin{adjustbox}{max width=\textwidth}

\begin{tabular}{|c|c|ccc|ccc|c|c|ccc|ccc|c|}

\cline{3-9} \cline{11-17}

\multicolumn{1}{c}{} &
\multicolumn{1}{c}{} &
\multicolumn{7}{|c|}{\textbf{One face image per template (1-1)}} &
\multicolumn{1}{c}{} &
\multicolumn{7}{|c|}{\textbf{Five face images per template (5-5)}}
\\  \cline{3-9} \cline{11-17}

 \multicolumn{1}{c}{} &  & \multicolumn{3}{|c|}{\textbf{Same-Pose}} &  \multicolumn{3}{c|}{\textbf{Cross-Pose}} & \textbf{Over-} &  & \multicolumn{3}{|c|}{\textbf{Same-Pose}} &  \multicolumn{3}{c|}{\textbf{Cross-Pose}} & \textbf{Over-} \\  \cline{1-1} \cline{3-8} \cline{11-16}

\textbf{Network} &  

& \textbf{F-F} & \textbf{3/4-3/4} & \textbf{P-P} 

& \textbf{F-3/4} & \textbf{F-P}  & \textbf{3/4-P}  & \textbf{all}

&  & \textbf{F-F} & \textbf{3/4-3/4} & \textbf{P-P} 

& \textbf{F-3/4} & \textbf{F-P} & \textbf{3/4-P}  & \textbf{all}

\\ \hhline{=~=======~=======}

\textbf{SqueezerFaceNet (ours)}  &  &

5.32 & 4.87 & 7.36

& 5.09 & 7.32 & 6.47 & 6.07

&

& 0.27 & 0.3 & 0.85	

& 0.23 & 0.74 & 0.75 & 0.52

\\ \cline{1-1} \cline{3-9} \cline{11-17}

SqueezeFacePoseNet 
\cite{[Alonso20SqueezeFacePoseNet]} &  & 
6.39 & 5.47 & 7.88 

&  6.09 & 8.15 & 7.02 & 6.34   

&  

& 0.27 & 0.06 & 0.54  

&  0.2 & 1.23 & 0.88 & 0.52 

\\ \cline{1-1} \cline{3-9} \cline{11-17}

\multicolumn{15}{c}{} \\

\end{tabular}


\end{adjustbox}

\label{tab:results-baseline-nopruning}

\end{table}

\begin{table}[htb]

\caption{Number of biometric verification scores.}
\centering

\begin{adjustbox}{max width=\textwidth}


\begin{tabular}{|c|c|c|c|c|c|c|}


\multicolumn{1}{c}{} & \multicolumn{1}{c}{} & \multicolumn{2}{c}{\textbf{SAME-POSE}}  & \multicolumn{1}{c}{} & \multicolumn{2}{c}{\textbf{CROSS-POSE}}  \\  \cline{3-4} \cline{6-7}

\multicolumn{1}{c}{\textbf{Template}} &  \multicolumn{1}{c|}{} & \textbf{Genuine} & \textbf{Impostor} & \multicolumn{1}{c|}{} & \textbf{Genuine} & \textbf{Impostor} \\ \cline{1-1} \cline{3-4} \cline{6-7}

1 image (1-1) & \multicolumn{1}{c|}{} &  368 $\times$ (9+8+...+1) = 16560  & 368 $\times$ 100 = 36800 & \multicolumn{1}{c|}{} & 368 $\times$ 10 $\times$ 10 = 36800 &  368 $\times$ 100 = 36800 \\ \cline{1-1} \cline{3-4} \cline{6-7}

5 images (5-5) &  \multicolumn{1}{c|}{} & 368 $\times$ 1 = 368  & 368 $\times$ 100 = 36800 & \multicolumn{1}{c|}{} & 368 $\times$ 2 $\times$ 2 = 1472 &  368 $\times$ 100 = 36800 \\ \cline{1-1} \cline{3-4} \cline{6-7}


\end{tabular}



\end{adjustbox}

\label{tab:scores}

\end{table}

\begin{figure}[htb]
\centering
    \includegraphics[width=0.8\textwidth]{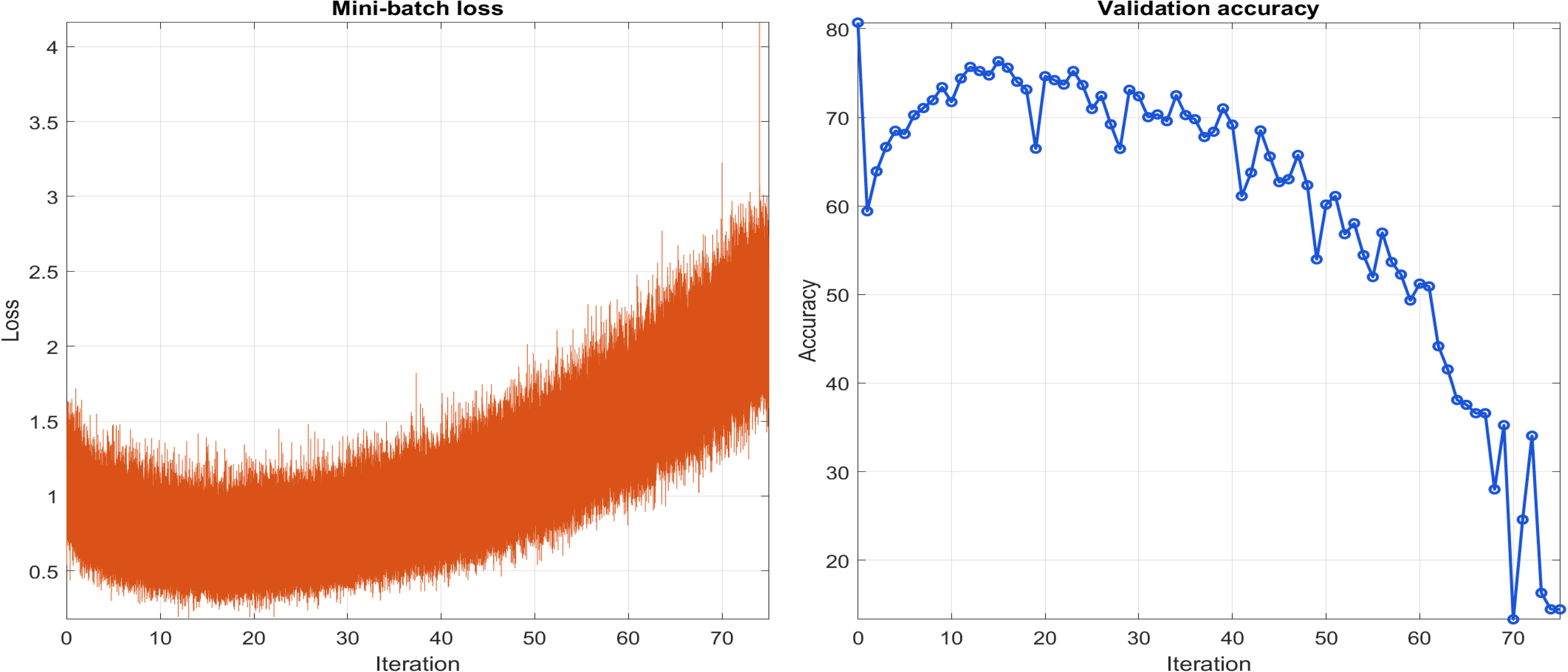}
    \caption{\label{fig:pruning-loss-accuracy}Mini-batch loss and validation accuracy during the pruning of SqueezerFaceNet. One iteration removes 1\% of the filters with the lowest importance scores.}
\end{figure}

\begin{figure}[htb]
\centering
    \includegraphics[width=0.8\textwidth]{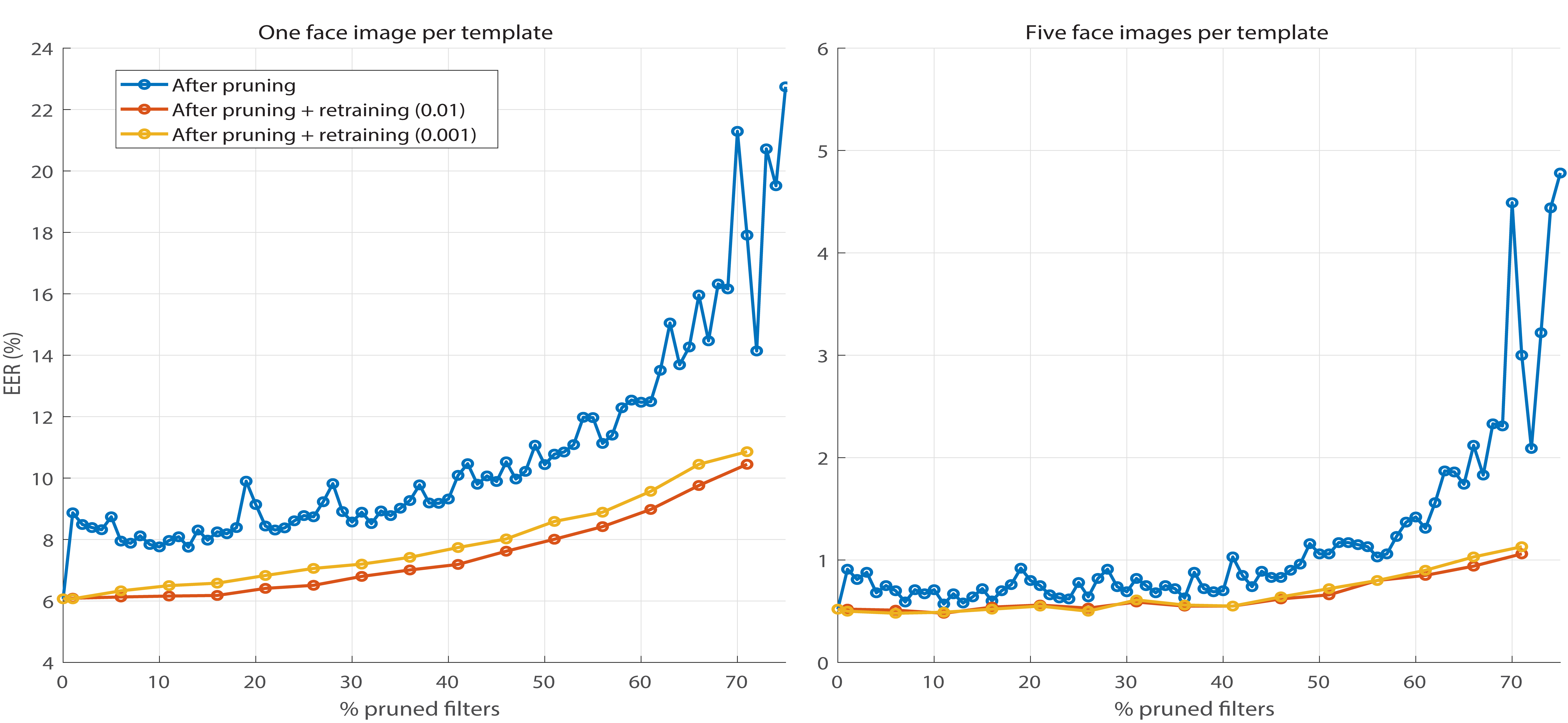}
    \caption{\label{fig:eer-all}Face verification results on the VGGFace2-Pose database (EER \%) during the pruning of SqueezerFaceNet (overall accuracy across all pose comparison types). One iteration removes 1\% of the filters with the lowest importance scores.}
\end{figure}

\section{Results}
\label{sect:result}

We first report the verification accuracy of SqueezerFaceNet without any pruning in Table~\ref{tab:results-baseline-nopruning}.
This will be the baseline to which we will compare after pruning.
We also give the results of SqueezeFacePoseNet from \cite{[Alonso20SqueezeFacePoseNet]}.
We detail the results of both same- and cross-pose experiments (Figure~\ref{fig:protocol}), as well as the overall performance across all poses. 
Same-pose comparisons are made with only templates generated with images of the same pose, while cross-pose experiments are done between templates of different poses. 
Genuine (mated) scores are obtained by comparing each template of a user to the remaining templates of the same user, avoiding symmetric comparisons. For impostor (non-mated) scores, the first template of a user is used as the enrolment template and compared with the second template of the next 100 users. 
Table~\ref{tab:scores} shows the total number of scores.

One observation from Table~\ref{tab:results-baseline-nopruning} is that SqueezerFaceNet improves the results of \cite{[Alonso20SqueezeFacePoseNet]}.
The main differences of the present paper 
are that we have added batch normalization to the network, we apply random horizontal flip to the training images, we use cosine similarity instead of $\chi^2$ distance to compare vectors, and we compute an image descriptor by averaging the descriptor of the original image and its horizontally flipped
counterpart \cite{[Duong19MobiFace]}.
These modifications seem to have an overall positive effect. 
Regarding pose comparison types, it can be seen that the worst performance is given by the most difficult ones, either when the image is only visible from one side (Profile vs. Profile) or when there is a maximum difference between query and test templates (Frontal vs. Profile).
It is also worth noting the substantial improvement observed when five images are used to generate user’s templates (5-5) in comparison to using one (1-1).

\begin{figure}[htb]
\centering
    \includegraphics[width=0.4\textwidth]{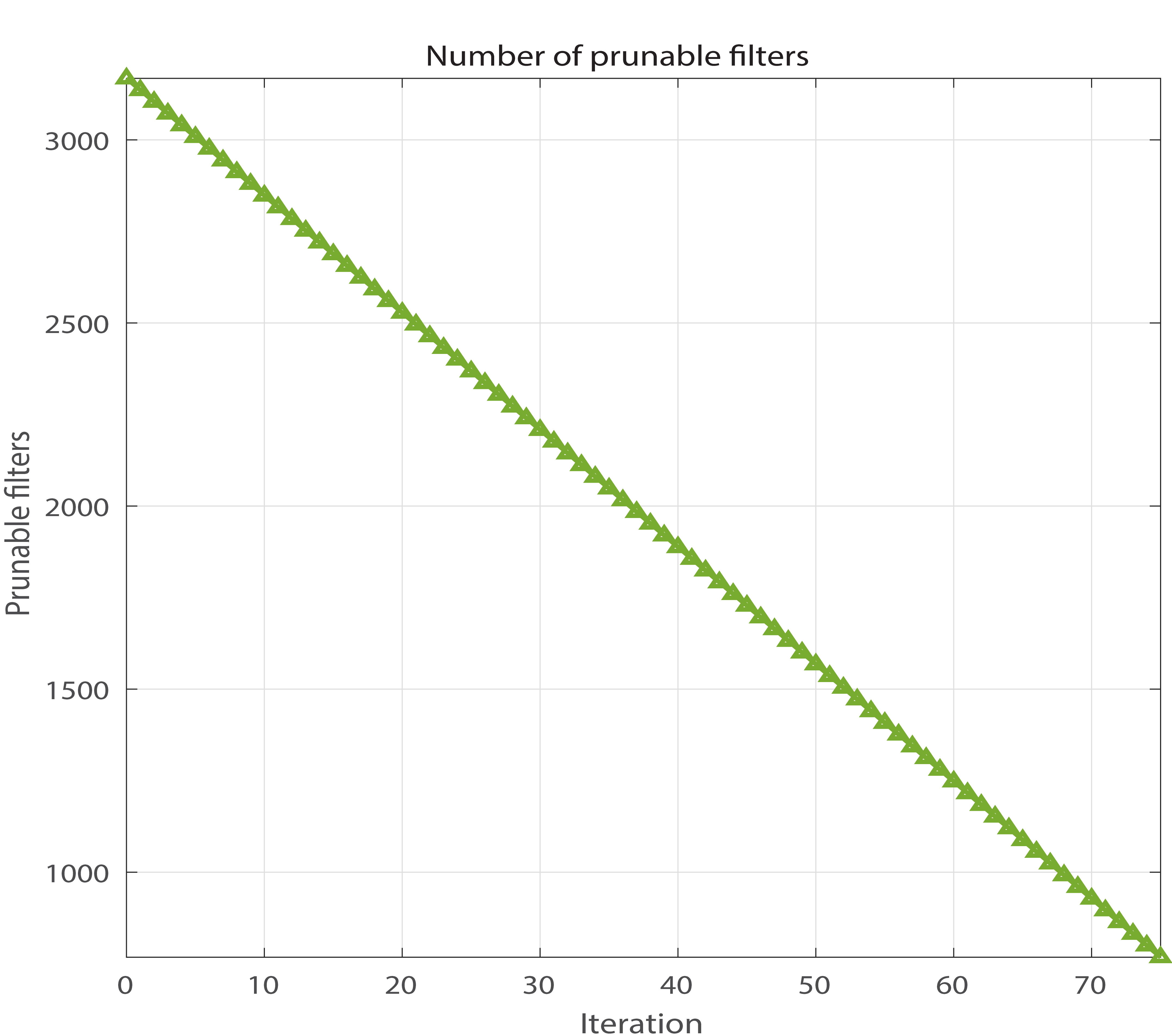}
    \includegraphics[width=0.4\textwidth]{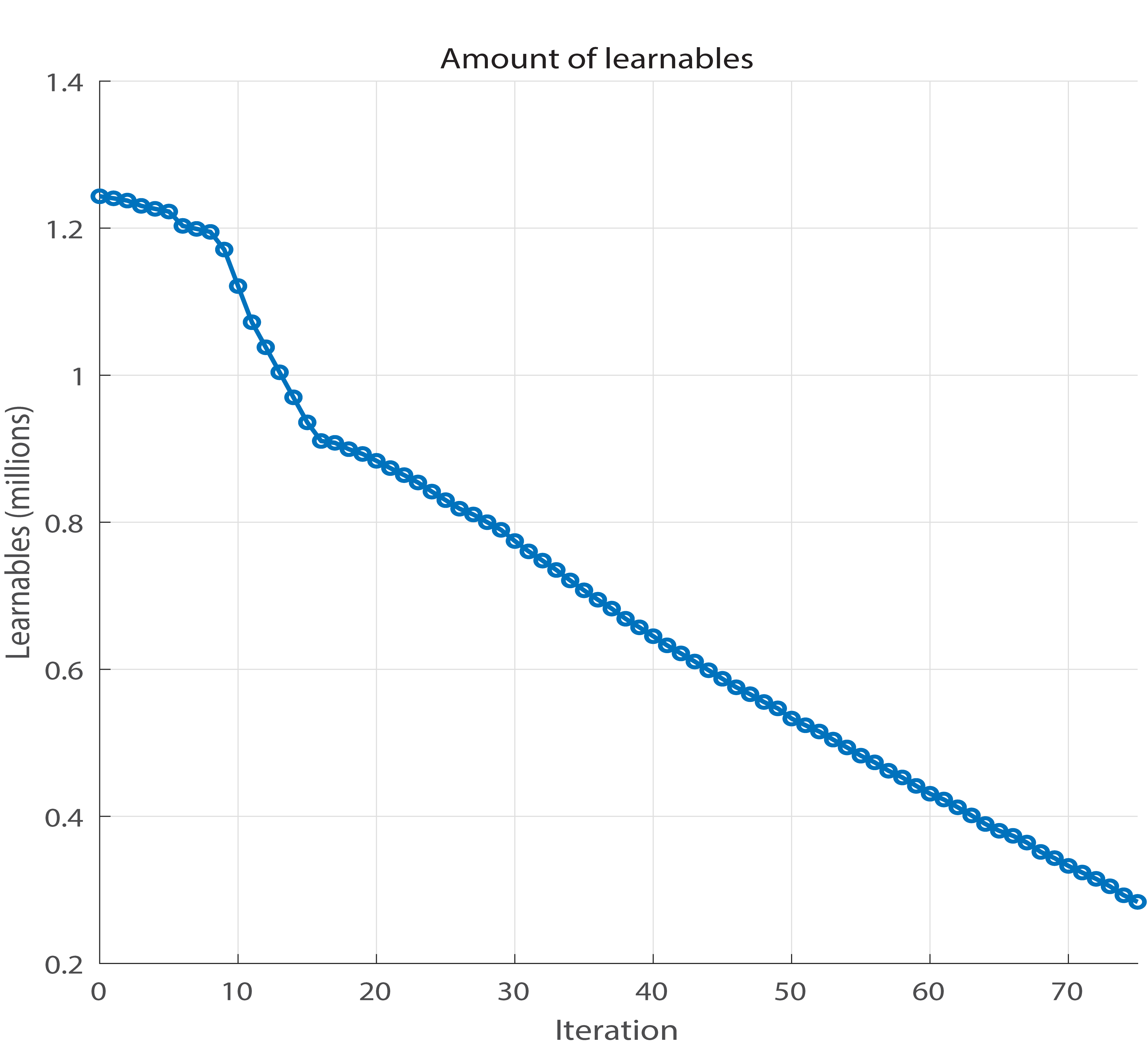}
%
    \includegraphics[width=0.4\textwidth]{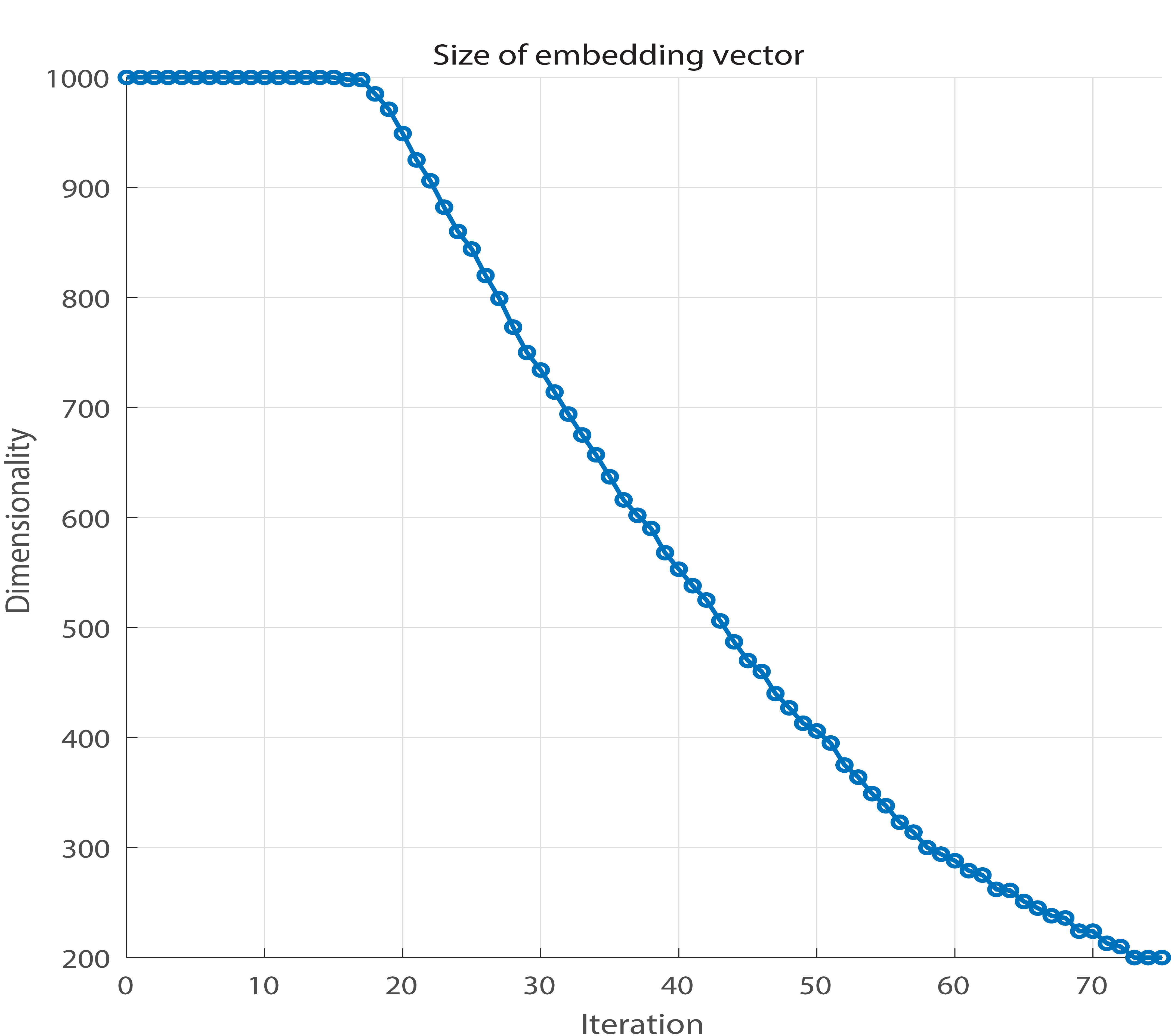}
    \includegraphics[width=0.4\textwidth]{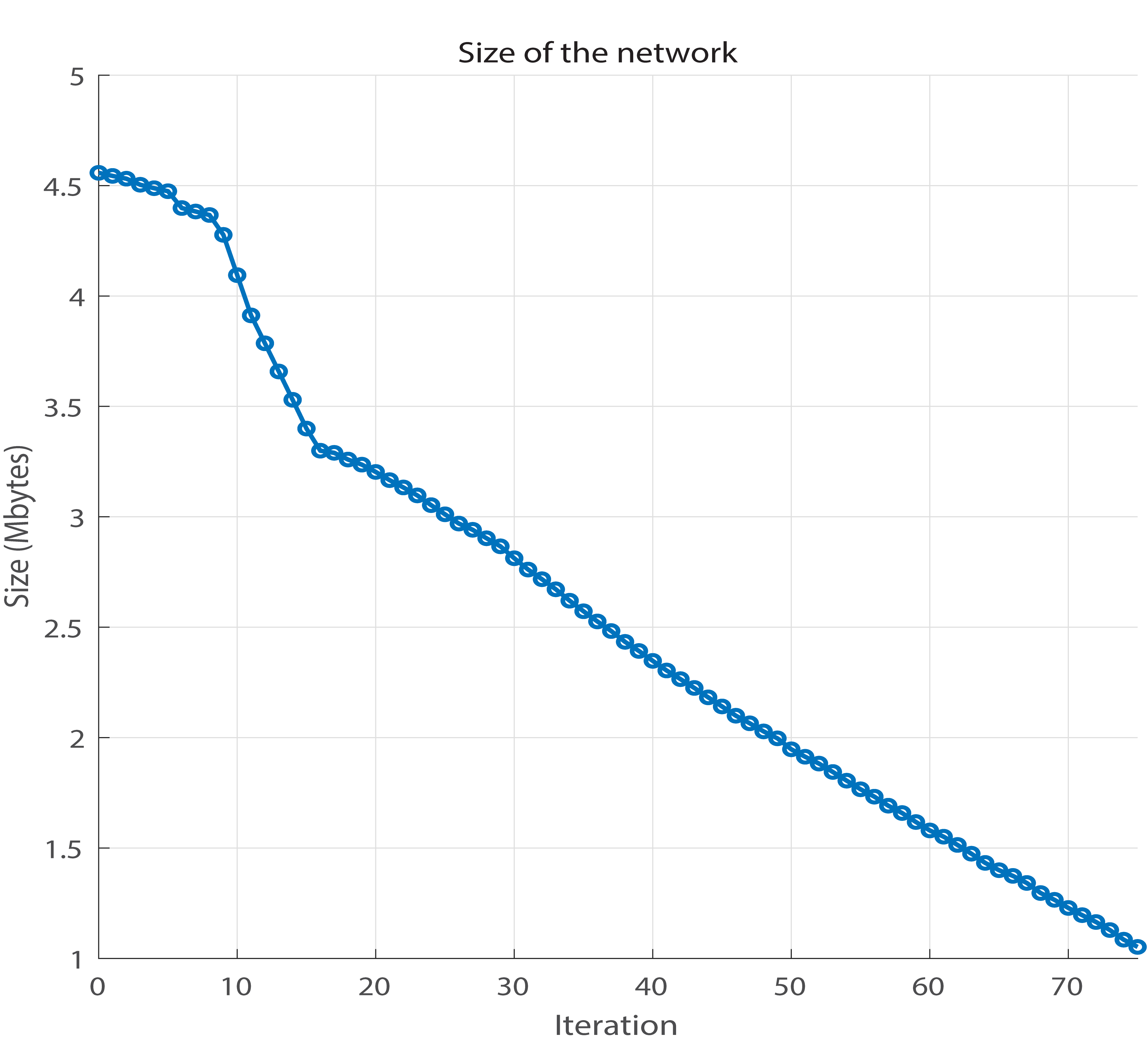}
    \caption{\label{fig:pruning-filters-learnables-embedding-file}Effect of pruning in: number of filters (top left), learnables (top right), embedding size (bottom left) and size (bottom right) of SqueezerFaceNet.}    
\end{figure}

We then apply the pruning of Sect.~\ref{sect:pruning_method} to SqueezerFaceNet.
On each iteration, we use a random 25\% of the VGGFace2 training set to compute the importance score of each convolution filter.
After each iteration, we remove 1\% of the filters with the lowest scores.
The optimizer is SGDM with a mini-batch of 128 and a learning rate of 0.01.
Figure~\ref{fig:pruning-loss-accuracy} shows the mini-batch loss and validation accuracy across different iterations.
An interesting observation is that the loss decreases a bit until $\sim$15\% of the filters have been pruned and then increases again (the validation loss shows the opposite behavior, as expected). However, after removing just 1\% of the filters (first iteration), the validation accuracy decreases sharply from $\sim$80\% to $\sim$60\%, and then it is regained again as the network is pruned up to $\sim$15\% of the filters. 
Figure~\ref{fig:eer-all} (blue curves) shows the overall verification accuracy of the pruned network on the VGGFace2-Pose database.
The origin of the $x$-axis ($x$=0) corresponds to SqueezerFaceNet without pruning. As can be seen also here, after removing just 1\% of the filters, there is a jump towards a worse performance, after which performance is regained a bit until the network is pruned approximately by 10-15\%. In five-to-five comparisons (right plot), performance is kept more stable until 30-40\% of the network has been pruned, suggesting that combining several face images to create a user template can be a method to counteract the effect of eliminating convolution filters. In one-to-one comparisons, however, accuracy decreases quicker.

\begin{figure}[htb]
\centering
    \includegraphics[width=0.8\textwidth]{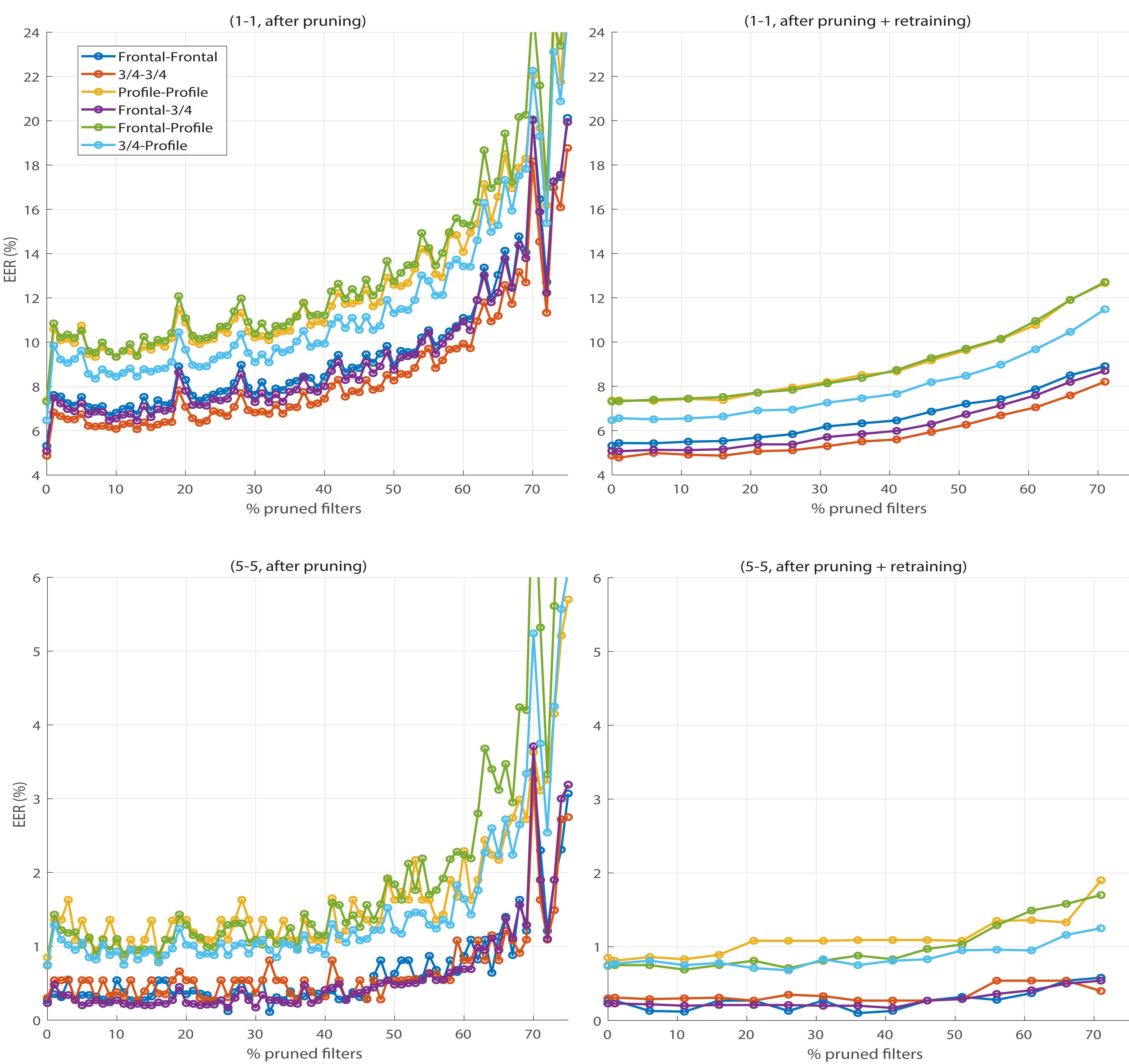}
    \caption{\label{fig:eer-per-pose}Face verification on VGGFace2-Pose (EER \%) during the pruning of SqueezerFaceNet per pose comparison type (retraining with a starting rate = 0.1). Each iteration removes 1\% of the filters with the lowest importance scores.}
\end{figure}

After pruning the network with different percentages, we retrain it over VGGFace2 according to the same protocol of the original unpruned network (Sect.~\ref{sect:protocol}) in order to regain the accuracy lost during pruning. 
Given the time that it takes to train the network over the entire VGGFace2, we do the retraining only every 5 iterations of the pruning algorithm (starting at 1\%). 
The results are given in Figure~\ref{fig:eer-all} as well. The network is retrained either with a starting learning rate of 0.01 (red curve) or 0.001 (orange). The rationale between these two options is that even if the network is pruned, it has already been trained once over the same database, so starting with a high learning rate may be counterproductive. However, as seen in Figure~\ref{fig:eer-all}, this is not the case. Indeed, in one-to-one comparisons, the best accuracy is given by starting with 0.01. Regarding the accuracy lost after pruning, it can be seen that training the pruned network again is able to recover the original accuracy up to a certain percentage of pruned filters. In one-to-one comparisons, performance remains stable until $\sim$15\% of the filters have been eliminated. Then, accuracy worsens exponentially. In five-to-five comparisons, on the other hand, performance remains at the same level as the unpruned network until about 30-40\%. A remarkable result, in any case, is that after 70\% of the filters have been removed, the EER is less than double, so a certain reduction in the number of filters does not translate to accuracy in the same proportion. In five-to-five comparisons, the EER goes from 0.52\% (unpruned network) to 1.06\% (network pruned at 71\%).

We then analyze the effect on the network of the pruning process (Figure~\ref{fig:pruning-filters-learnables-embedding-file}). 
Obviously, the number of filters decreases linearly on each iteration (by 1\%), since we have designed the experiments that way. 
However, the amount of learnables or the size of the network first decreases slowly until about 10\% of the filters are removed. Between 10 and 15\%, there is a significant drop in learnables, and then the decrease is stabilized again at a slower pace. 
This suggests that the filters that are removed first are not big and/or do not affect a high amount of channels, but then, the pruning algorithm removes filters having a larger amount of parameters. 
Regarding the size of the embedding vector, it is maintained constant until a pruning of about 18\%, indicating that the filters that are removed first do not affect the last layer of the network.
If we set 15\% as the optimal pruning (from Figure~\ref{fig:eer-all}), it translates to a reduction in parameters from 1.24M to 0.94M (by 24\%) and in size from 4.6MB to 3.4MB (by 26\%). This is without losing accuracy significantly. In five-to-five comparisons, we could go even higher and prune about 40\% of the network, resulting in 0.65M parameters and 2.35MB (a reduction of 48\% and 49\%, respectively).

We finally give the verification accuracy per pose comparison type after network pruning (Figure~\ref{fig:eer-per-pose}), with and without retraining. Obviously, the same accuracy gains after retraining are also observed here, and how the performance is maintained until a certain percentage of the filters is removed. It is also more evident the oscillations per iteration when SqueezerFaceNet is pruned but not retrained (left column), an effect that is alleviated after retraining (right column). 
Table~\ref{tab:results-comparison-pruning} also details the exact per-pose EER values for different degrees of pruning.
It can be observed that the combinations that do not involve profile (P) images result in better performance. 
Still, in five-to-five comparisons, even the difficult profile-profile (P-P) or frontal-profile (F-P) comparisons provide a very competitive EER of 1\% or less.
The table also shows the results with two variants of ResNet50 deployed by the authors of the VGGFace2 database \cite{[Cao18vggface2]} having a much higher amount of parameters. They use input images of 224$\times$224 and produce a feature vector of 2048 elements.
These two networks clearly stand out in comparison to our SqueezeNet model but at the cost of a larger number of parameters and size ($\sim$150MB), which is infeasible for mobile applications.

\begin{table}[htb]

\caption{Face verification results of SqueezerFaceNet on the VGGFace2-Pose database (EER \%) with different degrees of pruning (pruned networks retrained with a starting rate = 0.01). 
F=Frontal View. 3/4= Three-Quarter. P=Profile.
Results with two large networks (ResNet50 variants \cite{[Cao18vggface2]}) are also shown.
}

\centering

\begin{adjustbox}{max width=\textwidth}

\begin{tabular}{|c|c|c|ccc|ccc|c|c|ccc|ccc|c|}

\cline{4-10} \cline{12-18}

\multicolumn{1}{c}{} &
\multicolumn{1}{c}{} &
\multicolumn{1}{c}{} &
\multicolumn{7}{|c|}{\textbf{One face image per template (1-1)}} &
\multicolumn{1}{c}{} &
\multicolumn{7}{|c|}{\textbf{Five face images per template (5-5)}}
\\   \cline{4-10} \cline{12-18}

 \multicolumn{1}{c}{} & \multicolumn{1}{c}{} &  & \multicolumn{3}{|c|}{\textbf{Same-Pose}} &  \multicolumn{3}{c|}{\textbf{Cross-Pose}} & \textbf{Over-} &  & \multicolumn{3}{|c|}{\textbf{Same-Pose}} &  \multicolumn{3}{c|}{\textbf{Cross-Pose}} & \textbf{Over-} \\  \cline{1-2} \cline{4-9} \cline{12-17}

\textbf{Network} &  \textbf{Parameters} &

& \textbf{F-F} & \textbf{3/4-3/4} & \textbf{P-P} 

& \textbf{F-3/4} & \textbf{F-P}  & \textbf{3/4-P}  & \textbf{all}

&  & \textbf{F-F} & \textbf{3/4-3/4} & \textbf{P-P} 

& \textbf{F-3/4} & \textbf{F-P} & \textbf{3/4-P}  & \textbf{all}

\\ \hhline{==~=======~=======}







No pruning  & 1.24M &  &

5.32 & 4.87 & 7.36

& 5.09 & 7.32 & 6.47 & 6.07

&

& 0.27 & 0.3 & 0.85	

& 0.23 & 0.74 & 0.75 & 0.52

\\ \cline{1-2} \cline{4-10} \cline{12-18}

Pruning 16\%  & 0.91M &  &

5.53 &	4.87 &	7.38 &	5.16 &	7.52 &	6.64 &	6.18 &

&

0.27 &	0.31 &	0.89 &	0.21 &	0.75 &	0.78 &	0.54

\\ \cline{1-2} \cline{4-10} \cline{12-18}

Pruning 31\%  & 0.76M &  & 

6.19 &	5.3 &	8.2 &	5.71 &	8.13 &	7.26 &	6.80 &

 &

0.27 &	0.33 &	1.08 &	0.2 &	0.81 &	0.83 &	0.59

\\ \cline{1-2} \cline{4-10} \cline{12-18}

Pruning 46\%  & 0.58M &  & 

6.86 &	5.94 &	9.17 &	6.29 &	9.28 &	8.19 &	7.62 &

 &

0.27 &	0.27 &	1.09 &	0.27 &	0.97 &	0.83 &		0.62

\\ \cline{1-2} \cline{4-10} \cline{12-18}

Pruning 51\%  & 0.52M &  & 

7.21 &	6.27 &	9.63 &	6.74 &	9.7 &	8.48 &	8.01 &

 &

0.32 &	0.29 &	1.08 &	0.29 &	1.03 &	0.95 &	0.66

\\ \hhline{==~=======~=======}

ResNet50ft \cite{[Cao18vggface2]} & 25.6M &  & 

4.14 & 3.13 & 5.16 & 3.68 & 4.99 & 4.25 & 4.23 & 
& 
0.01 & 0.02 & 0.27 & 0.07 & 0.14 & 0.14 & 0.11

\\ \cline{1-2} \cline{4-10} \cline{12-18}

SENet50ft \cite{[Cao18vggface2]} & 28.1M &  &

3.86 & 2.87 & 4.16 & 3.36 & 4.48 & 3.71 & 3.74 &
 &
0.02 & 0.02 & 0.2 & 0.07 & 0.14 & 0.2 & 0.12

\\ \cline{1-2} \cline{4-10} \cline{12-18}

\multicolumn{15}{c}{} \\

\end{tabular}


\end{adjustbox}

\label{tab:results-comparison-pruning}

\end{table}

\section{Conclusions}

This paper deals with the task of developing SqueezerFaceNet, a lightweight deep network architecture for mobile face recognition.
For such purpose, we apply a CNN pruning method based on Taylor scores which assigns an importance measure to each filter of a given network. Such importance metric is based on the impact on the error if the filter is removed, and it only requires the back-propagation gradient for its computation. 
The method starts with a network trained for the target task (here: face recognition). 
Then, it is iteratively pruned by removing filters with the smallest importance.
To regain potential accuracy losses, the pruned network is finally retrained again for the target task.
The method is applied to an already light model (1.24M parameters) based on a modified SqueezeNet architecture \cite{[Iandola16SqueezeNet]}.
As training sets, we use the large-scale MS-Celeb-1M (3.16M images, 35K identities) \cite{[Guo16_MSCeleb1M]} and VGGFace2 (3.31M images, 9.1K identities) \cite{[Cao18vggface2]} datasets.
We evaluate two verification scenarios, consisting of using a different number of images to create a user template. In one case, a template consists of five face images with the same pose, following the evaluation protocol of \cite{[Cao18vggface2]}.
In the second case, we consider the much more difficult case of only one image to generate a user template.
Different pose combinations between enrolment and query templates are tested too (Figure~\ref{fig:protocol}).

Our experiments show that the pruning method is able to further reduce the number of filters of SqueezerFaceNet without decreasing accuracy significantly. 
This is especially evident if we employ a sufficient number of images to create a user template (five in our experiments). In such case, the number of filters can be reduced up to 40\% without an appreciable accuracy loss.
In one-to-one comparisons, a more difficult case, a reduction of up to 15\% is also feasible.
The resulting network in each case has 0.65M and 0.94M parameters, respectively.
As future work, we are looking into evaluating the employed pruning method in more powerful CNN architectures which are widely used in face recognition, such as ResNet \cite{Deng19CVPR_ArcFace}. 
%
%
If the same effects as in the present paper are observed, it would allow to lower error rates in comparison to the ones obtained in this paper (Table~\ref{tab:results-comparison-pruning}), for a fraction of the size of such large networks.
We are also considering other alternatives for network compression to evaluate if they are capable of producing even more reductions in network size \cite{Rattani23ACCESS_OcularCNNPruningBenchmark}.

\subsubsection{Acknowledgements.}
This work was partly done while F. A.-F. was a visiting researcher at the University of the Balearic Islands.
F. A.-F., K. H.-D., and J. B. thank the
Swedish Research Council (VR) and the Swedish Innovation Agency (VINNOVA) for funding their research.
Author J. M. B. thanks the project EXPLAINING - "Project EXPLainable Artificial INtelligence systems for health and well-beING", under Spanish national projects funding (PID2019-104829RA-I00/AEI/10.13039/501100011033).
We gratefully acknowledge
the support of NVIDIA Corporation with the donation of the Titan V GPU used for this research.
The data handling in Sweden was enabled by the National Academic Infrastructure for Supercomputing in Sweden (NAISS).

%
%
%
\bibliographystyle{splncs04}
%


\end{document}